\documentclass{article}

\usepackage[preprint]{neurips_2025}

\usepackage[utf8]{inputenc} 
\usepackage[T1]{fontenc}    
\usepackage{hyperref}       
\usepackage{url}            
\usepackage{booktabs}       
\usepackage{amsfonts}       
\usepackage{nicefrac}       
\usepackage{microtype}      
\usepackage{xcolor}         

\usepackage{graphicx}
\usepackage{subcaption}
\usepackage{wrapfig}
\usepackage{amsmath}
\usepackage{amssymb}
\usepackage{mathtools}
\def\equal{\texttt{=}}

\title{Structured Relational Representations}

\author{%
    Arun Kumar \\
    Dept. of Computer Science \\
    University of Minnesota, Twin Cities \\
    Minneapolis, MN, USA \\
    \texttt{kumar250@umn.edu} \\
    \And
    Paul Schrater \\
    Depts. of Computer Science and Psychology \\
    University of Minnesota, Twin Cities \\
    Minneapolis, MN, USA \\
    \texttt{schrater@umn.edu} \\
}

\begin{document}

\maketitle

\begin{abstract}
Invariant representations are core to representation learning, yet a central challenge remains: uncovering invariants that are stable and transferable without suppressing task-relevant signals. This raises fundamental questions, requiring further inquiry, about the appropriate level of abstraction at which such invariants should be defined and which aspects of a system they should characterize. Interpretation of the environment relies on abstract knowledge structures to make sense of the current state, which leads to interactions, essential drivers of learning and knowledge acquisition.  Interpretation operates at the level of higher-order relational knowledge; hence, we propose that invariant structures must be where knowledge resides, specifically as partitions defined by the closure of relational paths within an abstract knowledge space. These partitions serve as the core invariant representations, forming the structural substrate where knowledge is stored and learning occurs. On the other hand, inter-partition connectors enable the deployment of these knowledge partitions encoding task-relevant transitions. Thus, invariant partitions provide the foundational primitives of structured representation. We formalize the computational foundations for structured relational representations of the invariant partitions based on closed semiring, a relational algebraic structure.
\end{abstract}

\section{Introduction}
\label{sec:introduction}

Interactions with real-world objects are fundamental to the ability of both humans and artificial agents to perform tasks, acquire knowledge, learn, and make decisions. Agents must draw upon their existing knowledge to guide interactions, execute low-level plans, and revise their understanding of the environment when necessary. Such interactions may take various forms, including activities in the world, processing textual information, or visualizing and interpreting events or scenes. Interpretation underlies interaction; it arises from constructing both natural and abstract understandings of the environment that can be deployed as needed to make sense of the current state and plan subsequent actions. This interpretive ability relies on structured representations of relations and abstract knowledge, which are essential for learning, reasoning, and decision-making.

\begin{figure}[ht]
    \centering
        \begin{subfigure}[b]{0.25\linewidth}
            \centering
            \includegraphics[width=0.75\linewidth]{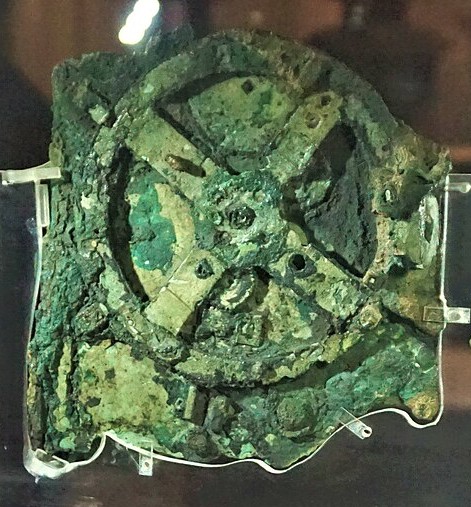}
            \caption{Interpretation}
        \end{subfigure}
        \begin{subfigure}[b]{0.365\linewidth}
            \centering
            \includegraphics[width=0.95\linewidth]{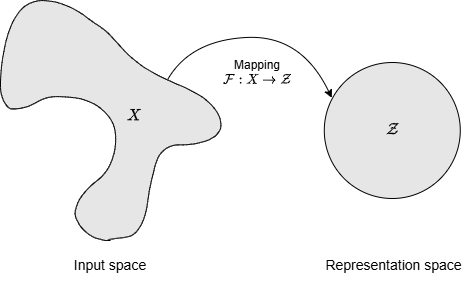}
            \caption{Representation learning}
        \end{subfigure}
        \begin{subfigure}[b]{0.365\linewidth}
            \centering
            \includegraphics[width=0.95\linewidth]{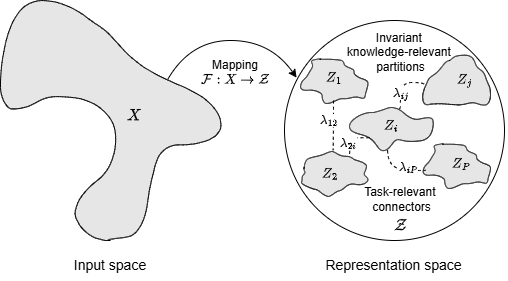}
            \caption{Structured relational representations}
        \end{subfigure}
    \caption{
    a) \emph{Interpretation} entails posing internal questions shaped by prior knowledge. For example, when interpreting an image of the Antikythera mechanism~\cite{antikythera_image}, individuals draw links from their prior knowledge.
    b) \emph{Representation learning} maps input to latent representation space by optimizing appropriate objective functions, with the aim of learning invariant representations while also retaining task relevant predictive utility. However, this core issue persists.
    c) \emph{Structured relational representations}, where invariant representations are at a higher knowledge level defined by partition blocks and the task relevant connectors, provide a promising offer to address core challenges in representation learning.
    }
    \label{fig:intro}
\end{figure}
The motivation for this problem arises from the process of interpretation. Consider the image in Figure~\ref{fig:intro}, which depicts a historical artifact, and contemplate how different individuals might interpret it. Their interpretations vary depending on prior knowledge — some may identify familiar components, while others may be drawn to questions about its composition, origin, lighting effects, or functional purpose. Is it a steering unit, a computational device, or something entirely different?
Even in everyday tasks, people interpret and contextualize the world before interacting with objects in it. These varying perspectives do not simply reflect differences in knowledge, but rather alternative views on a common subject. Yet, a common thread emerges: People strive for closure in understanding via interpretation grounded in internal abstract knowledge systems. This process drives abstract reasoning, supports learning and knowledge acquisition, and guides meaningful interactions with the world.
We argue that the pursuit of closure should guide the development of representations that bridge internal knowledge structures with interactions. Unlike arbitrarily imposed forms, representations via closure are grounded in interpretations. This perspective prompts several foundational questions: What are the goals, mechanisms, and representational structures that mediate an agent’s internal knowledge and its interactions? These inquiries are central to advancing our understanding of representation and learning.

Learning representations \cite{bengio2013representation} seek to map data points into a latent representation space by optimizing a suitable objective function. Models are typically trained on large-scale datasets in the hope of inducing invariant representations that remain stable under input variations while preserving predictive utility for downstream tasks. However, a persistent challenge remains: achieving a balance between capturing invariant structure and preserving discriminative, task-relevant information. These foundational challenges raise critical questions: at what level of abstraction should invariant representations be defined, and what specific aspects constitute such representations? Since the aim of invariant representations is to encode stable and transferable structures and agents must rely on abstract knowledge to guide interactions and execute low-level plans, we argue that invariant representations must reside where knowledge is. Natural objects inherently exhibit relational structures and whole–part relations, which must be preserved within coherent and structured representations. Closure operations enforce consistency across relational paths, promoting a coherent representational structure.

As a result, we propose that invariant representations in natural systems are structured partitions in an abstract knowledge space, formalized through the principle of closure. 
Structured representations must satisfy three core requirements: \textit{Chaining} refers to the ability to sequentially compose relations into composite relations within partitions and make connections between partitions. \textit{Choice} entails selecting among multiple viable alternatives through querying and task-dependent filtering. \textit{Closure} ensures and checks consistency within partitions, enforces conditions for relational chaining, and enables redirection to alternative paths when necessary. Consistent paths within partitions thus emerge as the fundamental invariant representations. The path-based invariant representation view focuses on what topological representations are changing, and this is where learning takes place. These partitions function as structural knowledge units, each with defined entry and exit points, enabling seamless reasoning within the partition. The disjoint partitions can then be flexibly composed through connectors to support task-specific reasoning.
This perspective simultaneously addresses the dual needs of stable representations and task-relevant flexibility. Intuitively, it is analogous to how people can reason in structured blocks: reasoning proceeds effortlessly within a partition block, while problem solving revolves around how to stitch together multiple such blocks.

Therefore, structured relational computations must be encapsulated within the primitives of structured representations—\textit{the three C’s of structured representations}: \textit{closure}, \textit{choice} and \textit{chaining}. In this article, we formalize the computational foundations for structured representations using a relational algebraic structure \textit{closed semiring}, which satisfies the essential properties required for representing and reasoning about relational structures in natural systems.

\subsubsection*{Related Work}
\label{sec:relatedwork}
Learning invariant representations is a foundational goal in machine learning. The aim is to map observations into compact representations that remain stable under input variations while retaining predictive utility across contexts. A broad range of approaches, such as representation learning \cite{bengio2013representation}, contrastive learning \cite{chen2020simple}, and domain-adversarial training \cite{ganin2016domain}, seek to encode invariances either explicitly or implicitly by embedding them in objective functions. However, a persistent challenge lies in balancing invariance with discriminability: representations that are invariant to spurious variations may also suppress task-relevant signals \cite{locatello2019challenging,ahuja2020invariant}. Domain generalization aims to address this challenge by training models that generalize effectively to unseen target domains \cite{wang2022generalizing}. Strategies include learning domain-invariant features to align source and target distributions \cite{muandet2013domain,li2018learning}, or directly mitigating domain-specific biases. However, inability to anticipate the nature of distributional shifts in target domains often results in models that overfit to the source domains and fail to capture the transferable structure \cite{gulrajani2020search}. The premise of domain generalization thus hinges on the assumptions about source-target relationships and the ability of models to acquire representations that are truly transferable. 

Relational reasoning can be categorized into three core types: induction, deduction, and transduction. Induction refers to generalizing from specific observed instances to broader patterns - supervised learning. Deduction involves deriving logical conclusions from general premises — logic or symbolic systems. Transduction infers new examples directly from known examples without explicitly learning a general rule — semi-supervised methods and graph neural networks. 
Many studies in learning disentangled representations \cite{esmaeili2019structured,higgins2018towards,locatello2019challenging} and others have focused mainly on disentangling statistically independent axes or transforming object properties, in contrast, our aim is to develop representations for higher-order relational knowledge that facilitate interpretations in meta-level knowledge learning \cite{kumar2024kix} in knowledge, interaction, and execution. 
Recent developments in neural architectures, including graph neural networks \cite{gilmer2017neural} and transformers \cite{vaswani2017attention}, show promise in relational reasoning by modeling interactions among entities \cite{battaglia2018relational} or inducing symbolic abstractions \cite{evans2018can}. Nevertheless, these models lack mechanisms to represent core relational operations - composition, coherence, inversion - limiting their ability in complex reasoning tasks.

A central issue concerns the level at which invariant representations are expected to operate: sensory, object, interaction, or abstract knowledge levels. Most contemporary models as above focus on low-level or object-level embedding spaces \cite{webb2023systematic}, which often fail to capture essential semantic or relational structure. Furthermore, defining and identifying the specific aspects on what constitutes invariant representations remains an open question, often requires strong prior knowledge or assumptions about the underlying data generation process, which may not always be available or accurate. These concerns ultimately revolve around the appropriate structure in representations and at what level invariant representations should be.
Unlike contemporary representation learning approaches that induce invariance at the object level, we propose that invariant representations reside in abstract knowledge spaces because agents rely on abstract knowledge to guide interactions and execute low-level plans. 
These invariant representations take the form of self-contained partitions within abstract spaces  defined by coherent internal relational structures, and transitions between partitions are mediated by connectors, enabling structured and task-relevant relational flow.  We derive these invariant structures by identifying representations that preserve knowledge coherence across the operations that access and modify knowledge across contexts, tasks and learning: representations that ensure closure over composing, chaining, evaluating, modifying relations.  To understand these criteria requires a minimal exposition of relation algebra and closure presented below. 

\section{Relational Algebraic Structures}
\label{sec:background}

Structures in relational reasoning have requirements of sequencing (composition as intersection or multiplication) and choices (composition as union or addition). Furthermore, the structure must include a natural notion of concluding or closure of reasoning. As a result, structured relational computations must jointly implement chaining, choice, and closure: distinct operations of relation algebraic coherence. 

Although tensor computations can be used to implement relational algebraic operations, several aspects of critical reasoning in relation computations require a highly structured implementation approach. In particular, relations can be viewed as operators, hypergraphs, or k-ary {\em typed} partial functions that have algebraic composition and transform requirements. Our implementation closely adheres to a tensor approach while enforcing these requirements. We treat relations as typed partial functions, using subscripts to indicate inputs and superscripts to indicate outputs. Thus, relations are reduced to simpler tensors by currying (partial evaluation), and using this strategy, we can represent a network of validly composed relations as a directed graph over curried relations. By structuring all inputs and outputs as (type, data) tuples, $[ R_{t_{i_1},x_{i_1}}^{t_{o_1},y_{o_1}}, R_{t_{i_2},x_{i_2}}^{t_{o_2},y_{o_2}} ,\cdots, R_{t_{i_n},x_{i_n}}^{t_{o_n},y_{o_n}} ]$, relational compatibility can be decided locally by matching tensor index.

Semirings provide a convenient computational framework for these operations.  They \cite{dolan2013fun,gondran2008graphs} are algebraic structures that have distributive addition and multiplication without the need for multiplicative inverses. 
Semirings ubiquitously occur in problems with two kinds of composition: parallel (choice) and serial (chaining).  Closed semirings have an additional operation of closure, an important algebraic operation in ensuring coherence structure in natural systems. 

A \textit{semiring} is an algebraic structure \( (S, \oplus, \otimes, 0, 1) \) consisting of a set \(S\) with two binary operations, \(\oplus\) and \(\otimes\), satisfying the following properties for any \(a, b, c \in S\):
\begin{itemize}
    \item commutativity: \(a \oplus b = b \oplus a\)
    \item associativity: \((a \oplus b) \oplus c = a \oplus (b \oplus c)\), \((a \otimes b) \otimes c = a \otimes (b \otimes c)\)
    \item identity: \(0 \oplus a = a\), \(1 \otimes a = a \otimes 1 = a\)
    \item \(\otimes\) distributes over \(\oplus\): \(a \otimes (b \oplus c) = (a \otimes b) \oplus (a \otimes c)\),  \((a \oplus b) \otimes c = (a \otimes c) \oplus (b \otimes c)\) 
    \item \(0\) is absorbing for \(\otimes\): \(0 \otimes a = a \otimes 0 = 0\) 
\end{itemize}

A \textit{closed semiring} is a semiring with an additional unary operation \( {}^{*} \) called closure \( (S, \oplus, \otimes, {}^{*} ,0, 1) \)
which satisfies
\( 
    a^* = 1 \oplus (a \otimes a^*)
          = 1 \oplus (a^* \otimes a)
\)
for all \(a \in S\).
\(a^*\) is intuitively the union of all chainings of \(a\).
The closure operator generalizes the concept of reflexive-transitive closure in relational semirings or path closure in graphs.

\textit{Relational closure}
A directed graph \(G = (V, E)\) is given by a set of nodes \(V\) and a set of edges \(E\). A relation \(r\) is an ordered pair \((u,v)\) denoted by \(u r v\), where \(u, v \in V\).
The graph has a finite number of nodes, \(n=|V|\), and a relation matrix \(R\).
Ownership relations are common in the natural world, and binary relational graphs can be represented by binary relational matrices \cite{givant2017relation,doumane2021graph} as
\(R[i,j] = 1  \text{ if } i r j, 0 \text{ otherwise} \).
If $u$ and $v$ are two nodes in a graph, then a path between them goes from $u$ to $v$ along the graph's edges, i.e. a path is composed of relations between the two nodes. If $u r v$ and $v r w$, then $u r w$, implying that the relation is transitive. 
Adding relation matrices is the union of edges, and multiplication of relation matrices is relational chaining. 
It enables us to create new composite relations from existing relations. As the graph has a finite number of nodes, the transitive closure of its relation matrix is the union of the first $n$ powers of R, $R^* = R \cup R^2 \cup ... \cup R^n = \cup_{i=1}^n R^i $. The reachability relation informs if a node $j$ is reachable from another node $i$ through any number of edges, therefore, it is relational closure. 
Reflexive transitive closure can be computed using \cite{floyd1962algorithm}.
\cite{lehmann1977algebraic} shows that the closure matrix can be broken down into block structures that are recursively computable \cite{kozen2012design}. These findings are significant because closure in binary matrices assures that relational reasoning 
has a coherent chaining structure and a termination point. 

\section{Structured Relational Computation}
\label{sec:theory}
\begin{figure}[ht]
    \centering
    \includegraphics[width=0.99\linewidth]{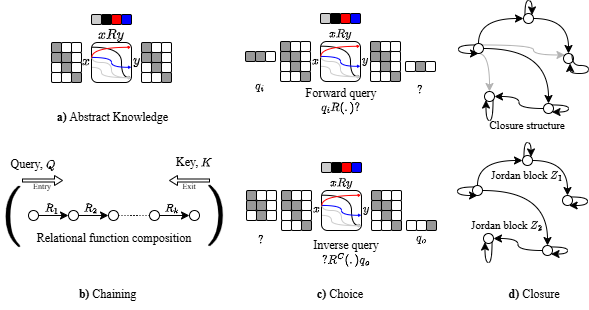}
    \caption{Overview of the core elements of the structured relational representations. Directed graphs are available after partial evaluation, as discussed above. a) Abstract knowledge is in the form of block partitions. In relational systems, it is not at object level but entity level relations, which are property to property relations. b) Relations can be chained together to form composite relations within block partitions, which define encapsulated units of abstraction and have an entry and exit points. c) Partitions can be queried for choices through forward or inverse query. Forward query results in which properties are related to inquiry while inverse query results in from which properties the answer can be arrived at. d) Closure illustrates reflexive transitive closure and block partition structures in a simple example. These three element together conform to the working of coherent and compositional systems.}
    \label{fig:structure}
\end{figure}

Structured representations are based on the notion of invariant path structures, which constitute the loci of knowledge. Interacting relational monoidal categorical structures \cite{cranch2024interacting} allow formalization of sequential and parallel composition, while the closure operation checks consistency. Partitions are partial relational paths projected onto a basis, and the computational aim is to uncover these basis paths, partitions, and connectors. In this section, we outline the core concepts and computations (see Figure \ref{fig:structure}) underpinning invariant structures-- chaining, choice, and closure.

\subsubsection*{Relational Invariant Representation Structures}
Knowledge structures emerge as directed relations between abstract characteristic features, as illustrated in Figure \ref{fig:structure}(a,d), and can be interpreted using relational paths. Invariant structures arise as partitions defined by closure operations. While closure ensures global coherence, analyzing partial order structural blocks enables us to identify which local relations are consistent with the knowledge. These partial paths represent partitions where relational knowledge is concrete, and localized queries yield consistent answers, even though reasoning requires relational chaining within a block. Thus, the abstraction of block structures under partial closure supports partial coherence in the relational flow.

Given a directed acyclic graph (DAG) $\mathcal{G} = (\mathcal{P}, \mathcal{E})$, the partially ordered set (poset) $(\mathcal{P}, \preceq)$ is defined by the directed relations $u r v$ where the relation connects node $u$ to $v$ in the graph and $v$ is relational successor of $u$, $u \preceq v$ in the graph. 
Given a finite poset $(\mathcal{P}, \preceq)$ of elements $\{x_1, x_2, \dots, x_n \}$, the zeta function $\zeta$ is defined as $\zeta(x,y) \equal 1 \text{ if } x \preceq y \text{ and } 0  \text{ otherwise}$ for all $x, y \in \mathcal{P}$.  If $f: \mathcal{P} \rightarrow \mathbb{R}$ is a function on the poset, then the zeta transform of $f$ is given by $ F(y) \equal \sum\limits_{x \preceq y} f(x)$ and in matrix notation, the zeta transform can be written as $F \equal Zf$ where $Z$ the zeta matrix is an $n \times n$ matrix  $ Z_{ij} \equal 1 \text{ if } x_i \preceq x_j, 0 \text{ otherwise} $ where $F$ is the transformed function. The M{\"o}bius function $\mu$ of a poset $\mathcal{P}$ is defined recursively as $\mu(x,y) \equal
-\sum\limits_{x \preceq z \prec y} \mu(x,z)  \text{ if } x \prec y \text{ and }
1 \text{ if  } x \equal y $. The M{\"o}bius transform recovers the original function $f$ from the transformed function and is given by $ f(y) \equal \sum\limits_{x \preceq y} \mu(x,y) F(x)$ and in matrix form $f \equal MF$ where $M$ is the M{\"o}bius matrix given by $M[i,j] \equal \mu(x_i, x_j)$.  
Directed graphs define poset structure by chaining relations; thus, the zeta transform functions as a partial relational closure operator.  

In a poset, elements are partially ordered through a directed relation, where the ending node is the successor of the relation's starting node. When two elements or nodes are not comparable under the given partial order and so have no chaining relation, they form an antichain.  Formally, an \textit{antichain} is a subset of a poset such that any two elements in the subset are incomparable under the partial order, which in case of poset induced by a directed graph means a subset of nodes such that no two nodes in the subset can reach another. Given a post $(\mathcal{P}, \preceq )$, a subset $ \mathcal{A} \subseteq \mathcal{P}$ is an antichain if $ x \not\preceq y \text{ and } y \not\preceq x  \forall x,y \in A$. 
The zeta matrix is the relational transitive closure on the poset of the DAG which can be computed through usual methods.  Antichains are identified from the transitive closure in a straightforward way by identifying nodes which are incomparable through relations.

\textit{Jordan blocks}
Given a scalar $\lambda$ and a positive integer $n \geq 1$, the \textit{Jordan block} $J_n(\lambda)$ of size $n$ associated with $\lambda$ is the $n \times n$ matrix defined as:
\[
J_n(\lambda) = 
\begin{bmatrix}
\lambda & 1 & 0 & \cdots & 0 \\
0 & \lambda & 1 & \ddots & \vdots \\
\vdots & \ddots & \ddots & \ddots & 0 \\
\vdots & & \ddots & \lambda & 1 \\
0 & \cdots & \cdots & 0 & \lambda
\end{bmatrix}
\]
where $\lambda$ appears on the diagonal, $1$'s appear on the super-diagonal just above the diagonal, and $0$'s elsewhere.
A square matrix $A^{n \times n}$ is said to be in \textit{Jordan normal form} if it is block-diagonal with Jordan blocks along the diagonal:
\[
J = 
\begin{bmatrix}
J_{n_1}(\lambda_1) & 0 & \cdots & 0 \\
0 & J_{n_2}(\lambda_2) & \cdots & 0 \\
\vdots & \vdots & \ddots & \vdots \\
0 & 0 & \cdots & J_{n_k}(\lambda_k)
\end{bmatrix}
\]
where each $J_{n_i}(\lambda_i)$ is a Jordan block corresponding to eigenvalue $\lambda_i$.
Each eigenvalue $\lambda_i$ may appear in multiple Jordan blocks and if there exists an invertible matrix $P$ such that $A = PJP^{-1}$, then $J$ is called a \textit{Jordan normal form} of $A$ and it is a canonical representative of the similarity equivalence class of A.
Each Jordan block $J_{n_i}(\lambda_i)$ in $J$ corresponds to a chain of generalized eigenvectors associated with eigenvalue $\lambda_i$. The set of all these chains forms a generalized eigenspace which is invariant under A and nilpotent within each generalized eigenspace. Each block encodes its chain length.
For Jordan block on relation matrix of a DAG, each block corresponds to a chain of nodes and each block represents paths. A Jordan block of size $k$ would mean a path of length $k-1$ in the DAG. 
To keep the mathematical notation consistent with familiar Jordan block notations in literature, we use $J$ to represent them, and denote partition blocks using $\mathcal{Z}$, as is common in representation learning. Both notations are used interchangeably for partition blocks throughout the text.
Therefore, each block partitions the abstract space into internal blocks structures. 

These internal block structures constitute the invariants of abstract knowledge representation, and learning fundamentally involves the discovery and organization of such invariant internal structures. The difficulty of unlearning or altering established beliefs appears to stem from the substantial cognitive cost of recomputing the foundational basis of these blocks. Each block possesses an internal relational flow, an invariant pattern of relational chaining, while external flows reflect dynamics of the system. Thus, learning representations is not merely a process of fitting mappings to inputs, but rather one of uncovering these invariant structures that underpin knowledge; learning is in essence knowledge learning. The partitions, embodied as internal relational blocks, serve as the core constituents of knowledge and form the essential elements of reasoning and learning. 

\subsubsection*{Forward-backward Query}
Block structures function as the \textit{modus operandi} for querying current understanding, see Figure \ref{fig:structure}c. Forward querying asks where one can reach from a given state through relational composition, while inverse querying asks from which prior states one could have arrived at the current state and through which relations and properties. This bidirectional mechanism of relational querying, forward and inverse, is analogous to forward and inverse planning. It constrains relational transitions by emphasizing that, before progressing arbitrarily to a new relation or node through interaction, one must query the underlying structural blocks. 
The choice monoid serves as a concrete view selection operation as long as we restrict choices to be grounded in data, allowing arbitrary selection of views constrained by available data. In contrast, query operations target specific ranges but may yield null results depending on the contents.
These operations do not involve events and we can do all the relational chaining with them.
The forward-backward mechanism or duality of analysis via relational paths and synthesis through the convergence upon abstract block structures anchors both inference and action in a coherent representational substrate.

\subsubsection*{Chaining}

Internal blocks are conceptualized as a coherent partition of knowledge, therefore, once in a block, relations can be composed through chaining, owing to the internal consistency of the block’s relational structure. As illustrated in Figure \ref{fig:structure}b, it is analogous to parsing, where the boundaries of structural units are clearly delineated. Each block possesses \textit{privileged entry and exit conditions} that align with its internal coherence, enabling structured reasoning to be invoked contextually. Such mechanisms are routinely employed by people: internal reasoning episodes are activated upon the satisfaction of certain entry criteria, and conclusions are reached, or transitions are made once exit conditions are met. Entry queries initiate reasoning within the block, while exit points mark the transition out of the current mode of structured reasoning facilitating transitions to new representational contexts. 

\section{Discussion}
\label{sec:discussion}
\subsubsection*{Connectors and External Flow}
\begin{figure}[ht]
    \centering
    \includegraphics[width=0.9\linewidth]{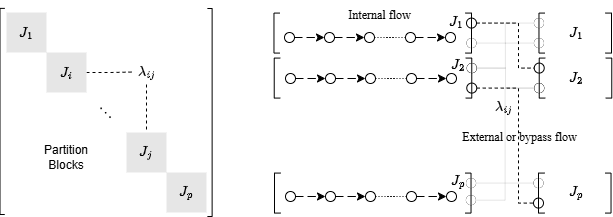}
    \caption{
    \textit{Connectors and external flow}: 
    Partition blocks capture internal relational structure and support coherent internal flows within which reasoning can proceed seamlessly. However, tasks necessitate deployment of these partitions and transitions across them. It introduces the need for connectors to facilitate external flows between partitions. Connectors serve as interfaces that enable the composition of otherwise disjoint partitions.
    }
    \label{fig:connectors}
\end{figure}
%
Systems can be conceptualized as dynamic flows of interaction, with block partitions being minimal view units. 
The block partitions function as interaction slots: within each block, relations can be coherently chained, facilitating seamless reasoning.
Importantly, graphs are not arbitrarily connected; their structure reflects preserved relational constraints between coherent partitions. This implies an underlying agreement in accessible constraints being preserved using a path view, which endows partition blocks with semantic meaning. From this perspective, constraints are naturally formalized as relational path dependencies, offering a principled approach for representing structured systems.

Leveraging these invariant partitions is therefore essential. In control theory, for instance, a typical approach involves selecting from a stack of control matrices under a set of constraints. In relational systems, constraints are relational path constraints on accessible properties. Knowledge thereby acts as a regulator, restricting permissible actions. At a low level, this translates to restrictions on the paths, while at a higher level, it emerges as capturing partial order relations, which can be modeled as abstract inequalities.
At the same time, interactions must also occur across partition boundaries. However, such external interactions require preconditions to be satisfied — this is the conceptual role of connectors or interfaces. An interface specifies the prerequisites to invoke a function or engage a partition block, as illustrated in Figure \ref{fig:connectors}.
In Jordan block decomposition, the columns of $P$ are the generalized eigenvectors, while each Jordan block in $J$ corresponds to a set of columns in \(P\) that form a Jordan chain.
These columns span a subspace related to the eigenvalues associated with that Jordan block.
Any external flow is seen through parameterizing a connector between two blocks with $\lambda_{ij}$, which can be determined by checking one point at a time $\delta_{ij}$. The $\lambda$'s facilitate querying what-if's and controllability of connector interfaces.
The compatibility of interfaces is essential for external interactions to occur, thereby defining a domain wherein partitions can be connected meaningfully. Partition blocks thus have the meaning of abstract entities in domain, while interfaces define boundaries for inter-block interactions.
Partition blocks therefore ensures internal coherence of relational flows, while connectors facilitate bypass flows across blocks. We often seek representations that can act as good translators and the invariant structures via partition blocks are well-suited.

\subsubsection*{Partition Block Structures}
\begin{figure}[!ht]
    \centering
    \includegraphics[width=0.99\linewidth]{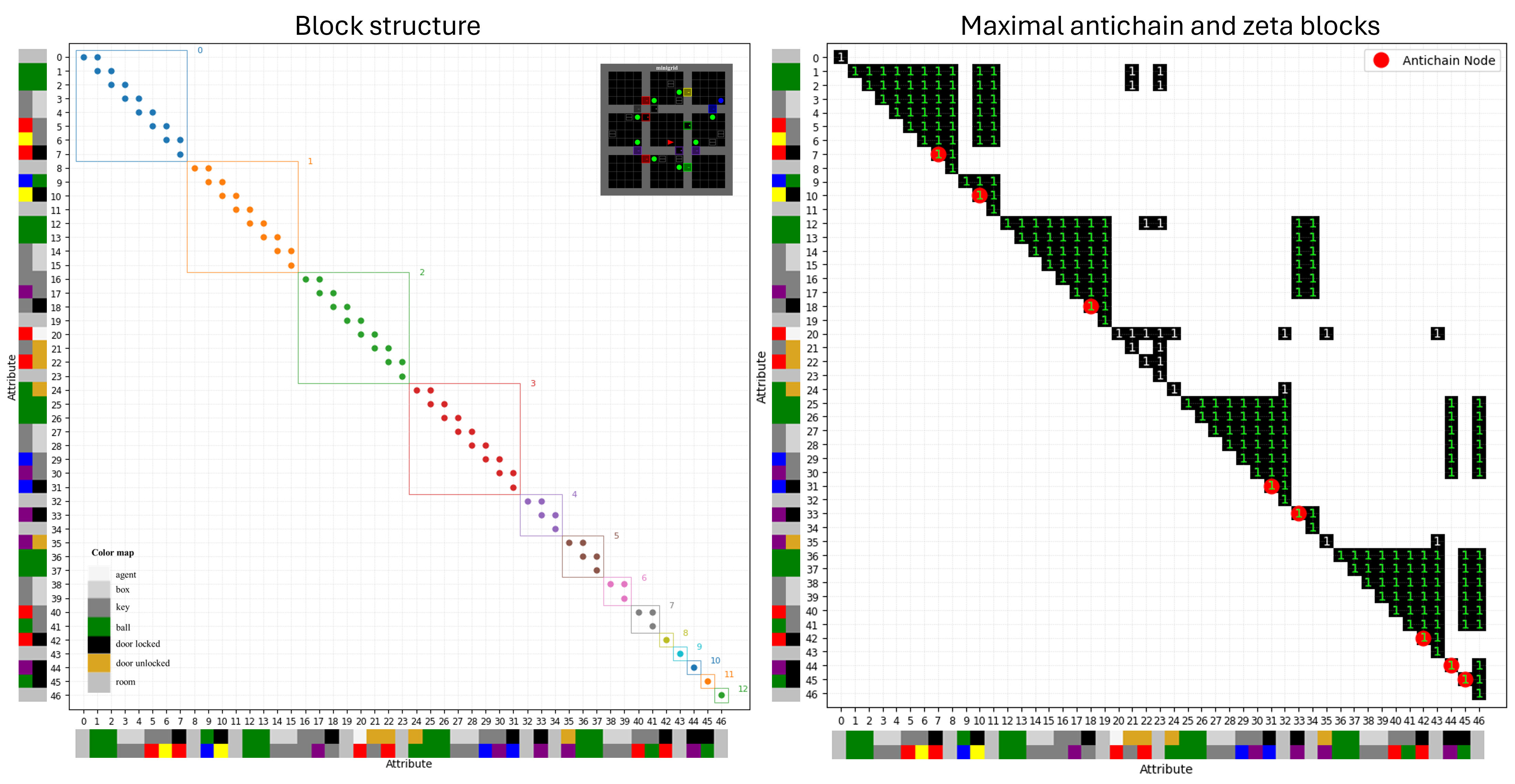}
    \caption{
    Visualization of partition block structures uncovered in application to feasible relations in a navigation environment (Inset). Axes are color-coded for clarity: the first row and column adjacent to the axes indicate the type, as defined by the accompanying colormap, while the second row and column denote the entity color.
    \textit{Left}: Jordan block partitions, where each block is distinctly highlighted. The source and sink nodes of each block are at the top and bottom, respectively.
    \textit{Right}: Maximal antichain nodes are marked with red circles while their corresponding zeta blocks over the reflexive transitive closure are highlighted in lime overlaying closure.
    }
    \label{fig:blocks}
\end{figure}

We uncover partition block structures, maximal antichains and the corresponding zeta blocks over the reflexive transitive closure, shown in Figure~\ref{fig:blocks}, by application to a feasible relation matrix in a navigation domain, the MiniGrid obstructed maze environment \cite{chevalier2018minimalistic}. 
The environment comprises nine rooms and objects such as balls, boxes, keys, and doors; some doors are locked requiring matching keys. The feasible relation matrix is precomputed, by encoding feasible binary relations between entity attributes. The considered relations are: `in\_room' - entity rooms; `is\_adj' and `can\_be\_moved' - movable entities balls, keys, and boxes which are either already adjacent or can be moved within the same room; `can\_contain' - a box can contain a key; and `open\_door' - a key can be used to open a door if it matches in color, or the agent can open an unlocked door.
The feasible relation matrix is an attributes-attributes binary relation matrix as described above. For the relation feasibility matrix, the Jordan blocks structure is revealed using SymPy. The relational reflexive transitive closure is computed using the Floyd-Warshall algorithm, while the maximal antichains are obtained by checking the reachability and then used to obtain the zeta block structures, see Figure~\ref{fig:blocks}.

Jordan blocks capture information about the intrinsic algebraic and geometric structure of the matrix. The structure of the uncovered blocks is shown in Figure \ref{fig:blocks} (Left).
The size of the largest Jordan block reflects the length of the longest chain, with each block possessing a distinct source and sink. In contrast, antichains consist of nodes that are mutually unreachable. A maximal antichain is the largest antichain. The maximal antichain and its associated zeta blocks are depicted in Figure \ref{fig:blocks} (Right).
Examining the maximal antichain nodes in conjunction with the Jordan block partitions reveals that maximal antichain nodes often interleave partition boundaries. The maximal antichain comprises the nodes \(\{7, 10, 18, 31, 33, 42, 44, 45\}\), and their corresponding zeta blocks over the reflexive transitive closure, both are illustrated in the figure. For instance, the Jordan block 1 spans from source node 8 to sink node 15, but the node 10 which is part of the maximal antichain lies within this partition as well. Similar overlap is observed for partitions 2 and 4, which include antichain nodes 18 and 33, respectively. In contrast, the remaining partitions of size greater than one have sink nodes, which coincide with a maximal antichain node.

While the Jordan blocks offer a partitioning mechanism based on algebraic structure, there will be partial paths that require transitions or connectors across blocks.
Antichain nodes, by contrast, are set of nodes that are mutually unreachable.
While reachability is given by the closure, antichains leads to nodes where one can connect into or out of partitions.
As visualized through the colormap, these antichain nodes correspond to locked doors in the environment.
Unlocking a locked door requires chaining through other nodes, such as retrieving a key from a box and then using it to open a door that matches key color.
In essence, analyzing block structure with maximal antichain indicates that antichains highlight the connectors where transitions occur to traverse between otherwise disjoint partition blocks. Together, the Jordan partitions and antichains appear to yield a more robust partitioning space, capturing both internal flows and the requisite connectors for block transitions.
%
\subsubsection*{Complexity Analysis}
Consider a domain that is not explicitly relational and consists of input, observation, abstract representation, and action spaces. Although the agent operates through actions, its reasoning is grounded in the abstract representation space. The partition structures serve as translators for guiding subsequent interactions, and the agent's decision-making is mediated via interpretation of these partitions. Partitions effectively constrain the state space, enabling localized reasoning within self-contained units. These partitions only need to be recomputed when the underlying knowledge itself changes, rather than requiring updates at every interaction. In this sense, knowledge serves as constraining the representation space into partitions, and transitions occur through partitions. Connectors further constrain the set of possible transitions between partitions, reducing the effective action space. 
%
In this context, let us briefly look at the impact of such partitions and connectors on forward planning in Markov Decision Processes (MDPs), which are widely applicable across domains involving diverse data modalities. 
Let $n$ be the number of states, $m$ the number of actions, $p$ the number of partitions. Assuming that $\epsilon$ is the fractions of partitions that require recompution once in a while. For the sake of simplicity, let the size of each partition be uniform $n_i = n/p$.  Jordan block partitions are precomputed in $\mathcal{O}(n^3)$. In standard MDP,  forward planning per step complexity is $\mathcal{O}(mn^2)$.
When the state space is organized into disjoint partitions, as hypothesized, the effective state space then becomes $S = \cup_{i=1}^{p} \mathcal{Z}_i; \mathcal{Z}_i \cap \mathcal{Z}_j = \emptyset$.
While it may resemble hierarchical or clustered MDPs, there is a key distinction: in typical hierarchical MDPs, transitions within partitions are often aggregated as distributions, which is a cost of $\mathcal{O}(mn^2)$. 
With the partition structure in typical MDPs, the estimated complexity for forward chains is $\mathcal{O}(\bar{m}p^2)$ by assuming that $\bar{m}$ transitions occur only between partitions.  
Structured partitions are defined by coherent internal structure, and updates are triggered only when inconsistencies arise. When updates are necessary, only the affected partition is recomputed, unlike aggregating the transition matrix. 
The Inter-partition flow is mediated by connectors with precondition and postcondition which further restricts the action space, let $c$ be the number of connectors. The time complexity is estimated to be $\mathcal{O}(cp^2)$. 
For a block $n_i$, recomputation cost is estimated $\mathcal{O}(\epsilon p n_i^3)$, while recomputation cost for partition MDP would depend on the dynamics model and could be significantly large due to aggregation.
Therefore, even in non-relational domains, the use of invariant block partitions and connectors offers computational advantages due to stable partitions and sparse connector transitions.

This work focuses on developing foundations of invariant knowledge partitions that support knowledge learning and representations; low-level navigation lies beyond the scope of this article.

\section{Conclusion}
\label{sec:conclusion}
Learning invariant representations, which is a foundational objective in machine learning, seeks to construct compact representations that remain stable under variations in input while retaining predictive ability across contexts. Despite significant progress, contemporary models face persistent challenges in uncovering invariants that are not only stable and transferable but also semantically meaningful and task-relevant. This raises fundamental questions about the appropriate level of abstraction at which invariant representations should be constituted and the specific aspects that define and characterize invariant representations.

We argued that invariant structures must reside where knowledge itself resides, specifically in the form of partitions defined by the closure of relational paths within an abstract knowledge space. A key aspect of invariant structure arises from interpretation in knowledge where closure between interaction and knowledge functions as a form of translation. Invariant representations should not be anchored at the object level but rather at a higher relational knowledge level, where knowledge is structured. These representations are best understood as partitions characterized by coherent internal relational structure, typically exhibiting a partial order, and with defined entry and exit points. The disjoint nature of these partitions allows independent engagement with them, while inter-partition relational flow necessitates connectors, a mechanism that enable transitions across partitions. Learning representations therefore becomes the process of discovering these invariant partitions, whereas task-relevant planning involves identifying and deploying them via appropriate connectors. In this article, we proposed foundations for structured representations of invariant partitions, formalized using a closed semiring and the principles of closure, choice, and chaining.

\bibliographystyle{named}
\bibliography{representation}
\end{document}